\pdfoutput=1

\documentclass[11pt]{article}

\usepackage[preprint]{acl}

\usepackage{times}
\usepackage{latexsym}
\usepackage{amsmath}
\usepackage{amssymb}
\usepackage{lmodern,babel,adjustbox,booktabs,multirow}

\usepackage[T1]{fontenc}

\usepackage[utf8]{inputenc}

\usepackage{microtype}

\usepackage{inconsolata}

\usepackage{graphicx}
\usepackage{lipsum}
\usepackage{kotex}
\title{PMoE: Progressive Mixture of Experts with Asymmetric Transformer
for Continual Learning}

\author{Min Jae Jung and  JooHee Kim \\
AI Lab, INFINIQ\\
  {\{mjjung, jhkim\}@infiniq.co.kr}}

\begin{document}
\maketitle
\begin{abstract}
Large Language Models (LLMs) encounter significant challenges in continual learning due to catastrophic forgetting, where new information overwrites previously acquired knowledge. This limitation leads to substantial environmental and economic waste. In this study, we introduce the PMoE, Progressive Mixture of Experts with Asymmetric Transformer, which aims to minimize forgetting by utilizing an asymmetric design with shallow layers dedicated to general knowledge and deep layers for new knowledge. 
PMoE incorporates progressively added experts in deep layers and a router that allocates new knowledge to the appropriate experts efficiently.
The router, positioned adjacent to the deep layers, utilizes deep features aggregating consolidated information. This enables the router to perform efficiently, allocating new knowledge to the appropriate experts, which progressively increase in the deep layers. Extensive experiments on TRACE datasets and general language understanding datasets demonstrate that the proposed PMoE outperforms previous state-of-the-art approaches.
\end{abstract}

\section{Introduction}

Although Large Language Models (LLMs) have advanced significantly \cite{gpt, llama},  unlike humans, they often forget previously acquired knowledge during continuous learning \cite{cl1, cl2, loramoe}. This phenomenon, known as catastrophic forgetting \cite{cl3}, results in substantial environmental and economic inefficiencies, particularly for generative LLMs that require extensive data for training according to scaling laws \cite{scalinglaw}. Consequently, research in continuous learning is crucial for the efficient recycling of language models. This paper focuses on developing techniques that preserve existing knowledge while simultaneously acquiring new information.

Existing continual learning approaches can be divided into three approaches: replay-based \cite{rep1, rep2}, regularization-based \cite{reg1}, and architecture-based \cite{para1, pp}. Replay-based and regularization-based methods use a memory buffer with examples from previous tasks or add regularization constraints to penalize changes in important weights. However, they still suffer from forgetting due to continuous adjustments in parameters. Recently proposed architecture-based approaches \cite{o-lora, pp} enhance performance in a parameter-efficient manner but still struggle with the trade-off between learning plasticity and memory stability \cite{cl_survey} or require prior information such as task-ID to classify input text into task-specific parameters.

In this study, we propose the Progressive Mixture of Experts with Asymmetric Transformer (PMoE), which efficiently adapts to new tasks while minimizing the forgetting of existing knowledge within a continual learning framework. The key idea of PMoE is its asymmetric depth design: shallow layers retain general knowledge, while deeper layers acquire new task-specific knowledge. To achieve this, PMoE progressively adds experts \cite{progressive} and employs a routing network to classify input text.

We take inspiration from \cite{labelword}, which shows that LLMs aggregate information in shallow layers around important words and distribute it in deep layers. We hypothesize that the performance of the router is improved when using features from deep layers rather than shallow layers. In PMoE, specialized experts are absent in shallow layers but present in deep layers, enabling the model to benefit from parameter efficiency and robust routing performance.

The main contributions of our work are summarized as follows:

\begin{itemize}
\item We propose PMoE, which efficiently adapts to new tasks for continual learning by introducing an asymmetric depth design.
\item Our method outperforms LoRA with replay-based and prior state-of-the-art methods on the TRACE benchmark.
\item Experimental results demonstrate that our asymmetric design is both effective and parameter-efficient in preserving prior knowledge while adapting to new knowledge.
\end{itemize}

\section{Background}
\subsection{Problem Statement}

In a continual learning setup, a model sequentially encounters datasets ${\mathcal{D}_1, \ldots, \mathcal{D}_T}$, where each dataset $\mathcal{D}_t = {(x^t, y^t)}$ consists of data points specific to task $t$. Concretely, the model is provided only with $\mathcal{D}_t$ during training for task $t$. The primary challenge in continual learning is to enable the model to retain previously acquired knowledge while effectively learning new tasks. In this study, we adopt the general setup of continual learning, which involves maximizing the cumulative performance across all tasks as follows:

\begin{equation}
\max_{\theta} \sum_{k=1}^T \sum_{{x,y} \in \mathcal{D}k} \log p_{\theta} (y | x)
\end{equation}

The overall performance ($OP$) and backward transfer score ($BWT$) are measured immediately following the completion of training for the $t$-th task, as follows:

\begin{equation}
\begin{split}
& OP_t = \frac{1}{t} \sum_{i=1}^t R_{t,i}^D \\
& BWT_t = \frac{1}{t} \sum_{i=1}^{t} (R_{t,i}^D - R_{i,i}^D)
\end{split}
\end{equation}

where $R_{t,i}^D$ refers to the score of the $i$-th task of the model after completing training on the $t$-th task ($i \le t$). Furthermore, we simulate a realistic setting by having the model generate text sequences without task ID at test time.

\subsection{LoRA}

Recently, many Parameter Efficient Tuning (PET) approaches have been proposed and widely adopted in fine-tuning settings. These methods aim to efficiently adapt pre-trained models by updating a smaller subset of parameters than updating all. Among these approaches, Low-Rank Adaptation (LoRA) \cite{lora} hypothesizes that weight updates in pre-trained models have a "low intrinsic dimension." Instead of updating the entire pre-trained weight $W_0 \in \mathbb{R}^{d \times k}$, LoRA updates decomposed bottleneck weights $B \in \mathbb{R}^{d \times r}$ and $A \in \mathbb{R}^{r \times k}$, where $W_0 + \Delta W = W_0 + BA$ and the rank $r \ll \min(d,k)$. Thus, the forward pass $h = W_0 x$ becomes

\begin{equation}
h = W_0 x + \Delta W x = W_0 x + BA x.
\end{equation}

In this paper, we utilize LoRA as the expert component of our model.

\section{Asymmetric Mixture-of-Experts}
\subsection{Overall Architecture}

\begin{figure*}[!t]
    \centering
    \includegraphics[width=1.55\columnwidth]{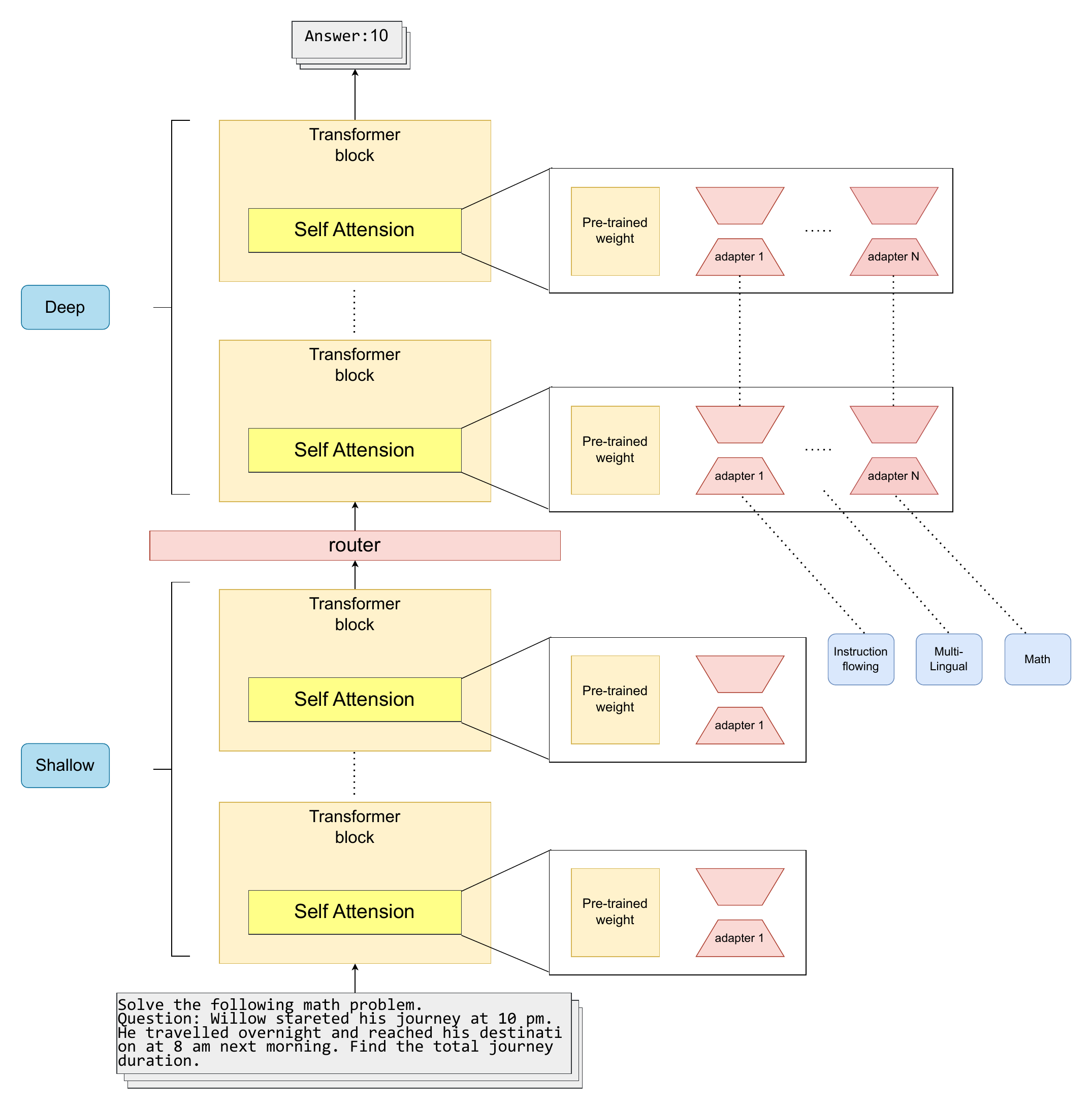}
    \caption{The overall architecture of PMoE for continual learning. Experts are located in each transformer block for fine-tuning, and the router is situated between the deep and shallow layers. Each shallow block contains only one expert, whereas deep blocks progressively add multiple experts as PMoE encounters new tasks.}
    \label{fig:our_model}
    \end{figure*}

To address the issue of information flow within transformer blocks, we propose an asymmetric Mixture-of-Experts (MoE) architecture as illustrated in Figure \ref{fig:our_model}. By introducing a threshold $\tau$, the overall network $f_\theta$ is divided into shallow layers $f_{\theta_{l=1,...,\tau}}$ and deep layers $f_{\theta_{l=(\tau+1),...,N}}$. The shallow transformer blocks employ a single LoRA $(BA)^l$ aligned with general human instruction-following, whereas deeper blocks utilize a multi-head LoRA $(BA)_k^l$ aligned with specialized abilities, assigning experts $k \in {1,...,T}$.

At the boundary between shallow and deep layers, the router distributes hidden features from the shallow layer to the deep layer. The routing network simply consists of a linear layer $W_g \in \mathbb{R}^{h \times T}$, and thus the distribution from routing $G(x) \in \mathbb{R}^{T}$ is represented as follows:

\begin{equation}
G(x) = \mathrm{Softmax} (x \cdot W_g).
\end{equation}

Many Mixture-of-Experts approaches adopt such routing networks to connect the input text to experts. However, there has been a major issue with the routers having fair connecting performance due to the overwhelming presence of a few experts, and various approaches have been proposed to overcome this drawback by introducing auxiliary loss \cite{sgmoe}, token capacity \cite{gshard}, or expert dropout \cite{sira}. Since PMoE uses deep features aggregating consolidated information, our asymmetric design does not require the application of such approaches. More details are provided in Section \ref{sec:routeranalysis}.

The forward pass of the self-attention layer can be expressed as:

\begin{equation}
f_{\theta_{l+1}}(h_l)=
\begin{cases}
W^l \cdot h_{l}, & \mbox{if} \ l \le \tau \\
W'^l \cdot h_{l}, & \mbox{otherwise}
\end{cases}
\end{equation}

where $W^l = W_0^l + (BA)^l$ and\\ $W'^l = W_0^l + \sum_{k=1}^T G(h_\tau)_k \cdot (BA)^l_k$.

To progressively allocate all experts with consolidated information, we progressively add the LoRA along with incremental tasks. This parameterization-based method avoids catastrophic forgetting by preserving the knowledge acquired from previous tasks.

\subsection{Experimental Setup}
\subsubsection{TRACE Dataset for Continual Learning Evaluation}

To evaluate fine-tuning as a scheme for continual learning, we adopt the TRACE dataset \cite{TRACE}, which comprises eight distinct datasets spanning multiple domains, including multilingual (C-STANCE \cite{c-stance}, 20Minuten \cite{20minuten}), code completion (Py150 \cite{py150}), mathematical reasoning (NumGLUE \cite{numglue}), and domain-specific (FOMC \cite{fomc}, MeetingBank \cite{meetingbank}, ScienceQA \cite{scienceqa}) datasets. Following Wang et al., we use a random sample of 5,000 training examples for each task, resulting in a total of 40,000 training examples, with the order always fixed as follows: C-STANCE, FOMC, MeetingBank, Py150, ScienceQA, NumGLUE-cs, NumGLUE-ds, 20Minuten.

\subsubsection{General Ability Evaluation}

To assess the impact of domain-specific tasks on the generality performance of pre-trained models, we evaluate five well-known benchmark datasets: MMLU \cite{mmlu}, GSM \cite{gsm}, BBH \cite{bbh}, BoolQ \cite{boolq}, and PiQA \cite{piqa}. Following the experimental setup of previous work \cite{TRACE}, we measure the general ability delta, which represents the performance difference between pre-continual learning and post-continual learning, as follows:

\begin{equation}
\Delta R^G_i = \frac{1}{M} \sum_{i=1}^{M}(R_{t,i}^G - R^G_{0,i}),
\label{eq
}
\end{equation}
where $M$ refers to the total number of benchmarks, and $R^G_{t,i}$ refers to the result on benchmark $i$ after the $t$-th tuning.

\subsubsection{Implementation Detail}
Our code implementation builds upon the PEFT \cite{peft} library. We employ Llama2-7b for our pre-trained weights and LoRA for our experts. We optimize our model using AdamW with a minibatch size of 128, a learning rate of 3e-4, and a cosine annealing scheduler. The hyper-parameter rank and $\tau$ are set to 4 and 24, respectively. At inference time, the model generates text sequences without knowing the task ID to which it belongs. All results are averaged over 3 runs.

Following the settings of previous studies \cite{TRACE}, we incorporate alignment data from LIMA \cite{lima} into the replay memory, replaying only 1\% of historical data.

\subsubsection{Baselines}
We compare PMoE with six baseline approaches. For consistency in experimental comparison, the foundation of all baselines is pre-trained Llama2-7b \cite{llama}. In Parameter Efficient Tuning (PET) comparisons, the number of trainable parameters is approximately consistent ($\le 0.1\% $).

\begin{itemize}
\item {{\bf Llama2-7b}* \cite{llama}: uses in-context learning with 6-shot on the current task without fine-tuning.}
\item {{\bf Full-ft} \cite{seqft}: fine-tunes full parameters on sequential tasks.}
\item {{\bf LoRA} \cite{lora}: fine-tunes fixed-size LoRA on sequential tasks.}
\item {{\bf O-LoRA} \cite{o-lora}: incrementally learns new tasks in a direction orthogonal to the LoRA subspace of past tasks while fixing the previous parameters.}
\item {{\bf Full-ft-RE}: fine-tunes full parameters with a memory buffer and replays samples from old tasks.}
\item {{\bf LoRA-RE}: fine-tunes fixed-size LoRA with a memory buffer and replays samples from old tasks, following the same memory setting as Full-ft-RE.}
\end{itemize}

\subsection{Main Result}

\begin{table*}
  \centering
  \begin{adjustbox}{max width=\textwidth}
{
  \begin{tabular}{l|c|c|c|c||c|c|c}
    \toprule
                 & \textbf{Llama2-7b$^*$} & \textbf{Full-ft$\dagger$} & \textbf{LoRA$\dagger$ } & \textbf{O-LoRA$\dagger$ }& \textbf{Full-ft-RE} & \textbf{LoRA-RE} & \textbf{PMoE} \\
    \hline
    MMLU {\small (5-shot)}& 46.6&46.4&42.3&46.7&43.4&41.9&46.8\\
    GSM {\small (8-shot)}&26.1&3.5&14.7&21.9&4.0&9.0&10.7\\
    BBH {\small (3-shot)}&40.2&30.1&33.1&40.3&30.9&31.7&37.1\\
    BoolQ {\small (0-shot)}&70.6&77.9&53.4&79.6&78.4&80.0&80.8\\
    PiQA {\small (0-shot)}&76.2&76.5&75.2&76.9&75.7&77.1&75.7\\
    \hline
    Average&51.9&46.9&43.7&53.0&46.5&47.9&50.1\\
    $\Delta R^G_t$&0&-5.0&-8.2&+1.1&-5.4&-4.0&-1.8\\
    \bottomrule
    C-STANCE&40& 45.4 & 27.7&48.2&49.5&48.1&47.5\\
    FOMC&48.3& 60.9 & 24&33.6&67.0&69.6&68.3\\
    MeetingBank&19.8& 45.7 & 12.1&40.9&47.5&43.4&38.5\\
    Py150&52.2& 51.2 & 0.4&53&55.1&53.3&55.1\\
    ScienceQA&62.8& 63.7 & 0&58.2&75.5&70.1&72.4\\
    NumGLUE-cm&28.4& 27.2 & 0&23.5&26.7&38.7&41.6\\
    NumGLUE-ds&20.3& 54.8 & 0&45.5&49.2&50.5&59.2\\
    20Minuten&39.5& 40.8 & 37&26.4&21.9&25.9&26.1\\
    \midrule
    Average&38.9&48.7&12.7& 41.2&49.0&49.3&51.1\\
    BWT &-&-8.3&-45.7&-6.2&-2.3 & +7.5 &+12.2\\

    \bottomrule
  \end{tabular}
  }
  \end{adjustbox}
  \caption{\label{citation-guide}
Comparison of the performance of PMoE and baselines. The upper part of the table indicates general language understanding and reasoning abilities, and the lower part of the table indicates the performance of continuously extended tasks after training on all tasks.
* denotes using in-context learning from 6 examples and $\dagger$ denotes results from TRACE \cite{TRACE}. All results are averaged over 3 runs.
  }
  \label{tab:main}
\end{table*}

The comparative analysis is presented in Table \ref{tab:main}. The upper part of the table represents the general capabilities of models that have undergone continuous learning to completion, while the lower part represents the specialized performance of models using the evaluation set from the data employed in continual learning. Furthermore, the left side of the table presents models that utilize in-context learning and those without the replay-based approach, whereas the right side presents models that incorporate the replay-based method.

Utilizing replay generally enhances performance, even with only 1\% of examples. Replay-based approaches show a BWT greater than zero, indicating not only no forgetting of past knowledge but also an improvement in performance due to recently acquired knowledge and memory. The PMoE outperforms LoRA-RE in both general and tuned abilities by 2.2\% and 1.8\%, respectively, and even exceeds the full-fine-tuning method, which has more than 1,000 times the trainable parameters, by 3.6\% and 2.1\% in general and task abilities, respectively.

The current state-of-the-art model, O-LoRA, demonstrates optimal general ability at 53.0\%, indicating an improvement in previously acquired knowledge. However, its performance on new tasks is relatively lower at 41.2\%. Notably, in the FOMC task, which was initially tuned, the PMoE demonstrated a 34.7\% improvement. Consequently, the PMoE not only enhances the performance of tuned tasks but also retains a high level of generalization ability.

\section{Discussion}

\subsection{Asymmetric Design}

\begin{figure}[!h]
\centering
\includegraphics[width=1.\columnwidth]{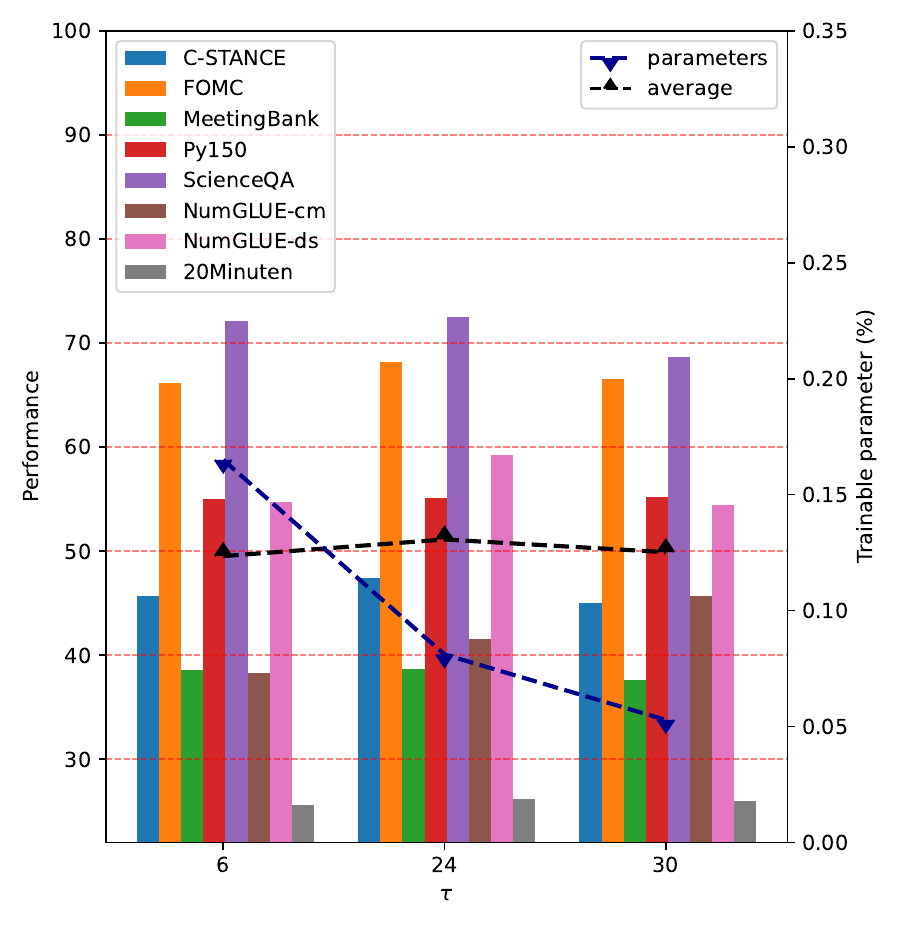}
\caption{The performance and computation according to shallow threshold $\tau$. The best performance is at $\tau=24$ and computation decreases in proportion to $\tau$.}
\label{fig:tau}
\end{figure}

The core strategy of the PMoE architecture is characterized by its asymmetric design, where shallow layers are dedicated to general abilities and deeper layers are tailored for newly acquired abilities. To assess the efficacy of this asymmetric design, we analyze the performance as functions of varying $\tau$. A lower $\tau$ results in the model possessing a greater number of task-specific parameters, with the routing mechanism utilizing shallow features. Conversely, a higher $\tau$ leads to a model enriched with parameters for general abilities, where the router utilizes deep features. As illustrated in Figure \ref{fig:tau}, the model demonstrates optimal average performance at a moderate $\tau$, and a decrease in computational cost proportional to $\tau$. Thus, if performance levels are comparable, it is advantageous to increase $\tau$ to reduce computational demands.

\subsection{Router Analysis}
\label{sec:routeranalysis}

\begin{figure}
\centering
\includegraphics[width=1.3\columnwidth]{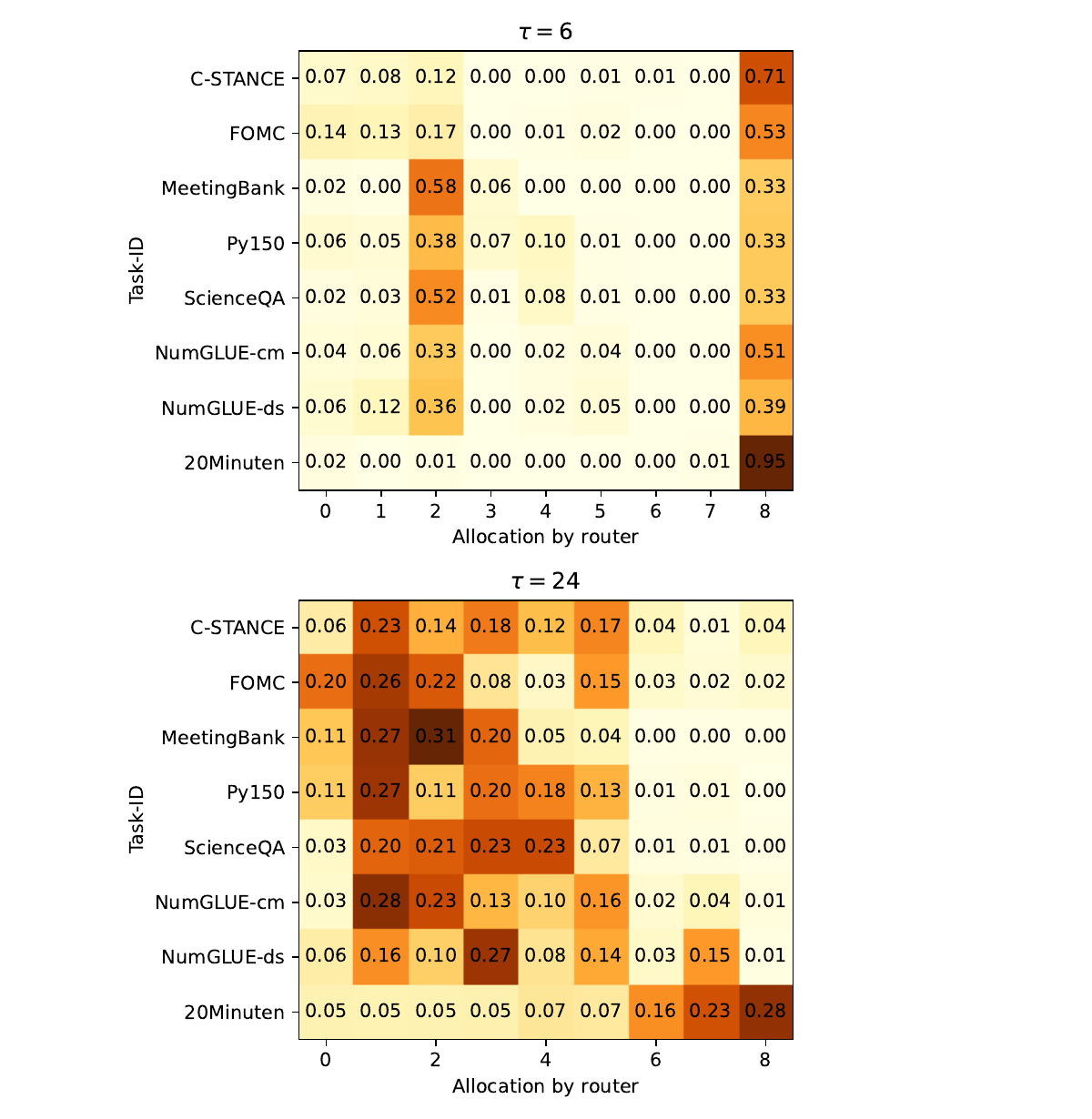}
\caption{Probability matrix in which a router allocates text from specific subsets to experts at (up) $\tau=6$ and (down) $\tau=24$.}
\label{fig:confusion}
\end{figure}

To investigate the router in PMoE, we show matrices representing the allocation probability by the router. As shown in Figure \ref{fig:confusion}, when the router is located near the shallow layer (i.e., $\tau=6$), the router mainly chooses only a few experts. This imbalance problem is a well-known phenomenon where input $x$ is assigned to a specific superior expert, rather than being distributed across a diverse range of experts. Previous works have attempted to address the router imbalance problem by introducing auxiliary losses \cite{sgmoe}, token capacity \cite{gshard}, or expert dropout \cite{sira}. However, when the router is located near the deep layer (i.e., $\tau=24$), the imbalance problem is automatically solved, well distributing all experts. As mentioned in the motivation of our study, this result supports the hypothesis that LLMs aggregate information in shallow layers and distribute it in deep layers, leading to better router performance when using features from deep layers rather than shallow layers.

\begin{figure*}[!h]
\centering
\includegraphics[width=1.95\columnwidth]{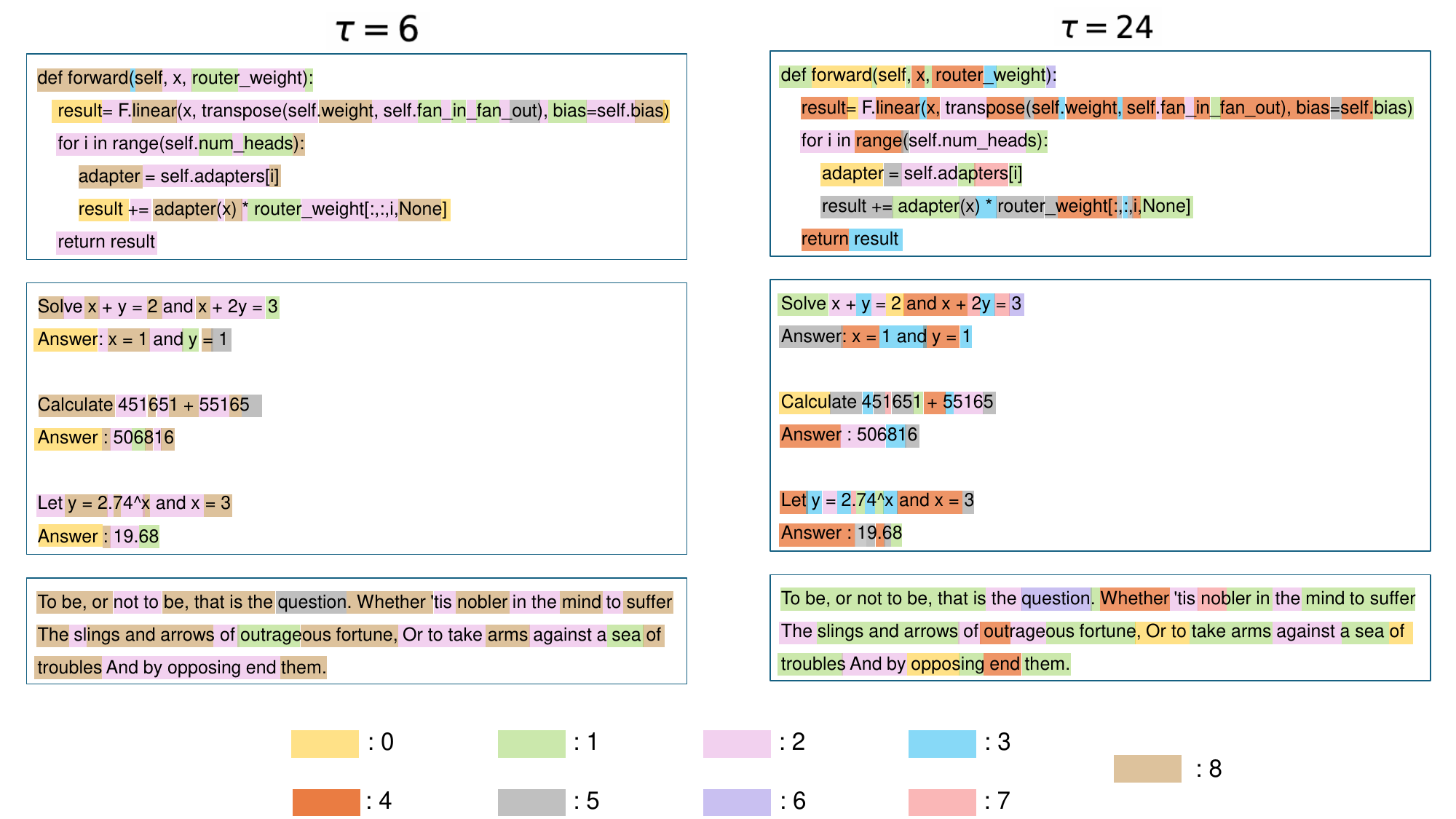}
\caption{The examples where each token is colored with the largest choice probability in the router at (left) $\tau=6$ and (right) $\tau=24$.}
\label{fig:qualitatively}
\end{figure*}

To further investigate, we conducted a qualitative evaluation of the token allocation by the router, as illustrated in Figure \ref{fig:qualitatively}. The evaluation utilized test prompts that include three distinct types: Python code, mathematical questions and answers, and lines from Shakespeare's Hamlet. With a $\tau$ of 6, most tokens are predominantly allocated to experts 2 or 8. In contrast, at $\tau$ of 24, the allocation patterns are more varied, indicating a comprehensive use of experts. Concretely, the router seems to allocate tokens to experts based on identifiable patterns related to the role of the tokens. For example, special characters are predominantly assigned to expert 3, elements of Python syntax are mainly assigned to expert 4, and plain text is typically assigned to expert 1.

Drawing from the aforementioned analysis, if $\tau$ is 24 and the task-ID is treated as the ground truth for the router during the training stage, PMoE is capable of employing each expert as a task-specific expert without the need for task-ID during the inference stage. We incorporate cross-entropy loss to guide the routing of input $x$ to a suitable expert, as follows:

\begin{equation}
\mathcal{L}(x, k) = - \log G_k(x)
\label{eq:auxloss}
\end{equation}

where $G_k(x)$ denotes the router probability of assigning input $x$ to its corresponding task-ID $k$. This loss provides a simple yet effective mechanism to prevent inputs from being pushed to a single expert by guiding the task assignment corresponding to input $x$.

\begin{figure}
\centering
\includegraphics[width=0.85\columnwidth]{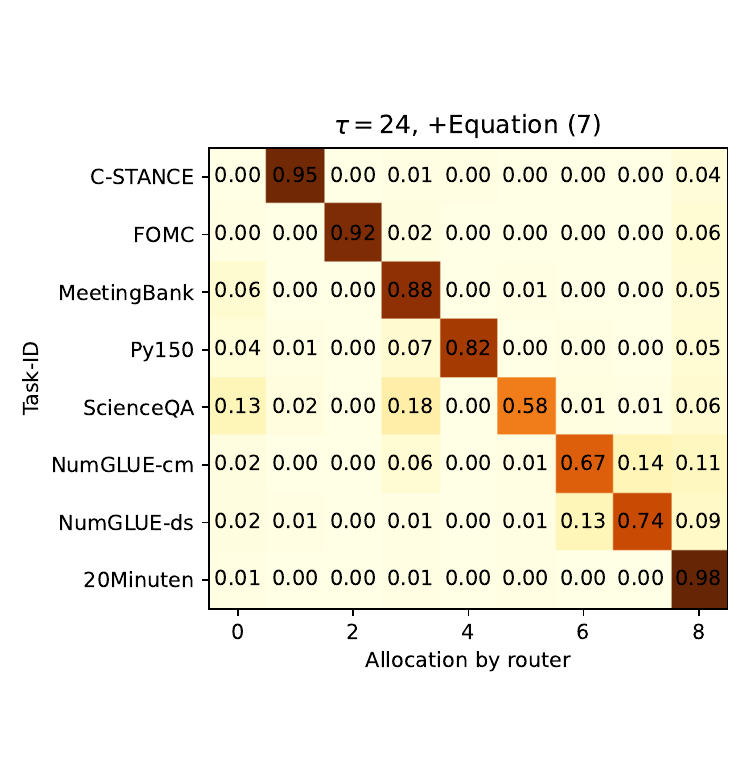}
\caption{Probability matrix in which a router with Equation \ref{eq:auxloss} allocates text from specific subsets to experts.}
\label{fig:auxloss}
\end{figure}

The comparison with the application of Equation \ref{eq:auxloss} is shown in Figure \ref{fig:auxloss}. This auxiliary loss function experimentally demonstrates its effectiveness in making experts task-specific without task-ID in the inference stage. However, in Table \ref{tab:auxloss}, applying the auxiliary loss slightly degrades overall performance. This result suggests that a combination of experts rather than task-specific experts has a positive effect on overall performance. Future research should explore extending task-specific experts for asymmetric design.

\begin{table}[!h]
  \centering
  \begin{adjustbox}{max width=\textwidth}
      {
  \begin{tabular}{l|c|c}
    \hline
     & {$R^G$} & {$OP$} \\
    \hline
    PMoE&50.1&51.1\\
    PMoE + Equation \ref{eq:auxloss} &49.4&50.4\\
    \hline
    \end{tabular}
    }
  \end{adjustbox}
  \caption{Performance comparison of PMoE with and without the auxiliary loss function.}
\label{tab:auxloss}
\end{table}

\subsection{Comparison with Trainable Parameter Size}
\begin{table*}
  \centering
  \begin{adjustbox}{max width=\textwidth}
      {
  \begin{tabular}{l|c|c|c|c|c|c|c|c|c|c|c}
    \toprule
     & {Params} &{\small C-STANCE} & {\small FOMC} & {\small MeetingBank} & {\small Py150}& {\small ScienceQA} & {\small NumGLUE-cm} & {\small NumGLUE-ds} & {\small 20Minuten}& $OP_t$ & {$R^G_t$}\\
    \hline
    LoRA-RE&0.077\% &48.1&69.6&43.4&53.3&70.1&38.7&50.5&25.9&49.3&47.9\\
    PMoE &0.081\% &47.5&68.3&38.5&55.1&72.4&41.6&59.2&26.1&51.1&50.1\\
    \midrule
    LoRA-RE&0.155\% &48.1&68.4&38.9&53.9&69.9&36.9&48.9&26.3&48.9&48.9\\
    PMoE &0.162\% &46.2&66.6&38.6&55.1&70.0&38.7&57.5&26.2&49.8&49.6\\
    \midrule
    LoRA-RE&0.310\% &48.1&68.0&39.5&54.2&70.8&39.9&51.3&26.2&49.8&48.8\\
    PMoE &0.324\% & 49.6&67.0&38.8&55.1&72.0&36.6&57.8&26.2&50.4&50.0\\
    \bottomrule
    \end{tabular}
    
    }
  \end{adjustbox}
\caption{Performance comparison of LoRA and PMoE for different parameter scales. In all cases, the pre-trained model is the same as Llama2-7b.}
\label{tab:parameter}
\end{table*}

We assessed the detailed comparison between PMoE and LoRA with the replay method at different scales of adaptation parameters, as shown in Table \ref{tab:parameter}. In our evaluation, we control the number of parameters by the rank of the adapter from under 0.1\% to over 0.3\%. In all parameter size settings, the general ability of PMoE is higher than that of LoRA by 0.7\%-2.2\%. In specialization performance, PMoE is higher than LoRA by 0.6\%-0.8\%. Interestingly, there is only a slight difference in performance along with the parameter size.

\section{Related Works}
\subsection{Parameter Efficient Fine Tuning}

As larger pre-trained models become more prevalent, efficiently fine-tuning their parameters has gained significant importance. Parameter Efficient Fine Tuning (PEFT) approaches are a set of methods that update only a small portion of parameters while freezing the pre-trained parameters during the fine-tuning stage. Various approaches have been proposed, including inserting additional modules or parameters, such as Adapter \cite{adapter}, Prompt Tuning \cite{prompt}, and Low-Rank Adaptation (LoRA) \cite{lora}.

In this paper, we focus primarily on LoRA, which hypothesizes that weight updates in pre-trained models have a "low intrinsic dimension." Building upon LoRA, AdaLoRA parameterizes incremental updates in the form of singular value decomposition \cite{adalora}, O-LoRA involves layering low-rank adapters on the key and value projection matrices \cite{o-lora}, SIRA leverages the Sparse Mixture of Experts to boost the performance of LoRA \cite{sira}, and PMoE employs an asymmetric design with shallow layers dedicated to general knowledge and deeper layers for new knowledge.

\subsection{Continual Learning}

Existing continual learning approaches can be categorized into three main types: replay-based, regularization-based, and architecture-based.

{\it Replay-based approaches}\cite{rep1, rep2} store examples from previous tasks for rehearsal. Although these approaches achieve strong performance, storing large samples can be inefficient and may not be feasible for data privacy settings.

{\it Regularization-based approaches} \cite{reg1, reg2} update only a few weights trained on previous tasks. These approaches are memory efficient but can suffer from catastrophic forgetting, especially in long sequence tasks.

{\it Architecture-based approaches} isolate weights trained on previous tasks and progressively add weights for new tasks. Although the recently proposed progressive prompt \cite{pp} has achieved state-of-the-art results, it has the limitation of requiring the task-ID during the inference stage. Among the continual learning techniques that do not require task-ID, O-LoRA \cite{o-lora} effectively addresses catastrophic forgetting. However, its performance, as measured by the TRACE benchmark \cite{TRACE}, was lower. In this paper, PMoE achieves superior performance on the TRACE benchmark without requiring task-ID during the inference stage.

\section{Limitations}

In our study, we conducted experiments across eight distinct tasks. Expanding the number of tasks could enhance the robustness and generalizability of our findings. Our research focuses solely on generative models, particularly Llama2-7b. However, the foundational concepts of our model hold potential for broader applicability in task-agnostic settings. Further research will assess the efficacy of our proposed approach across diverse benchmarks, aiming to broaden its application to multimodal continual learning.

\section{Conclusion}
In this paper, we enhanced the performance of LLM in continual learning by using PMoE. The PMoE features an asymmetric architecture between shallow and deep layers. Specifically, the shallow part employs a single expert to enhance general ability, whereas the deep part progressively utilizes an increasing number of experts to focus on tuned performance. Empirical experiments on comprehensive continual learning benchmarks demonstrate that the proposed PMoE outperforms previous state-of-the-art approaches. Furthermore, to investigate the asymmetric structure, we experimented by varying the asymmetric hyperparameter $\tau$, measuring the performance and efficiency of the model, and analyzing the router.

\bibliography{latex}

\end{document}